\documentclass[10pt,twocolumn,letterpaper]{article}

\usepackage{cvpr}              %
\newcommand{\ourmethod}{PerpetualWonder\xspace}

\newcommand{\myparagraph}[1]{\vspace{0.1cm}\noindent\textbf{#1}}
\renewcommand{\paragraph}[1]{\vspace{0.1cm}\noindent\textbf{#1}}

\usepackage{graphicx}

\newcommand{\myicongravity}{%
  \raisebox{-0.5ex}{\includegraphics[height=1.2em]{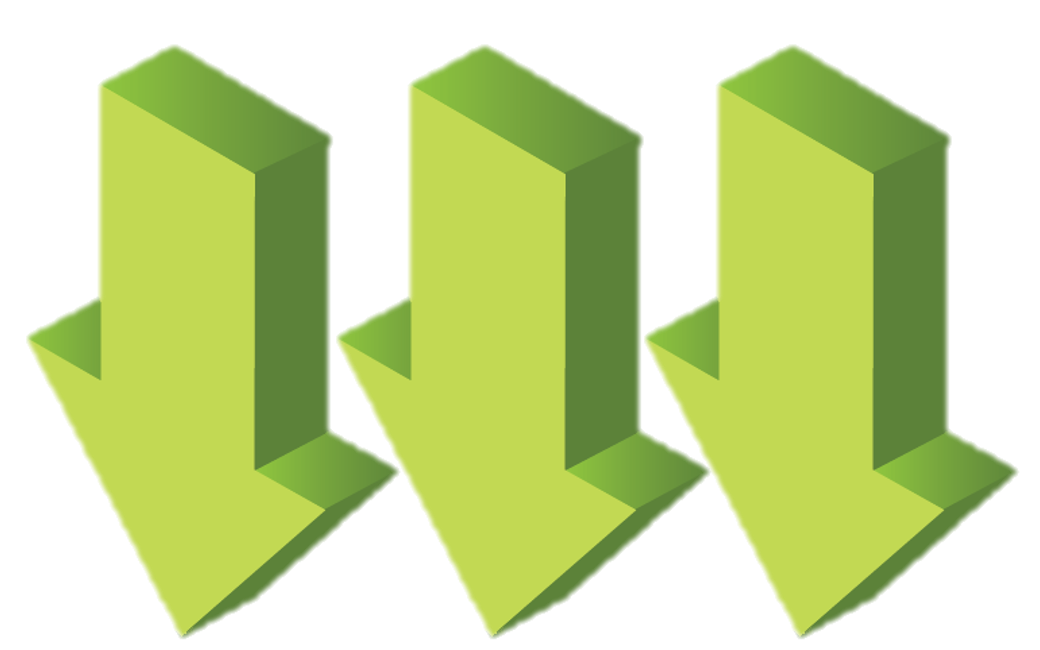}}%
}

\newcommand{\myiconforce}{%
  \raisebox{-0.5ex}{\includegraphics[height=1.2em]{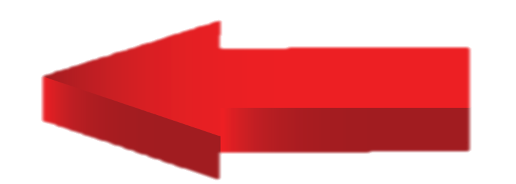}}%
}

\usepackage{mwe}

\usepackage[accsupp]{axessibility}  %

\definecolor{cvprblue}{rgb}{0.21,0.49,0.74}
\usepackage[pagebackref,breaklinks,colorlinks,allcolors=cvprblue]{hyperref}

\def\paperID{***} %
\def\confName{CVPR}
\def\confYear{2026}

\makeatletter
\newcommand{\printfnsymbol}[1]{%
    \textsuperscript{\@fnsymbol{#1}}%
}
\makeatother

\title{\ourmethod: Long-Horizon Action-Conditioned 4D Scene Generation}

\author{
Jiahao Zhan$^{*,\text{†}}$ \;
Zizhang Li$^*$ \; 
Hong-Xing Yu \;
Jiajun Wu \;
\\[0.5em]
Stanford University 
\\
\\
{\url{https://johnzhan2023.github.io/PerpetualWonder/}}
}

\begin{document}
\newcommand{\ba}{\boldsymbol{a}}\newcommand{\bA}{\boldsymbol{A}}
\newcommand{\bb}{\boldsymbol{b}}\newcommand{\bB}{\boldsymbol{B}}
\newcommand{\bc}{\boldsymbol{c}}\newcommand{\bC}{\boldsymbol{C}}
\newcommand{\bd}{\boldsymbol{d}}\newcommand{\bD}{\boldsymbol{D}}
\newcommand{\be}{\boldsymbol{e}}\newcommand{\bE}{\boldsymbol{E}}
\newcommand{\bff}{\boldsymbol{f}}\newcommand{\bF}{\boldsymbol{F}} %
\newcommand{\bg}{\boldsymbol{g}}\newcommand{\bG}{\boldsymbol{G}}
\newcommand{\bh}{\boldsymbol{h}}\newcommand{\bH}{\boldsymbol{H}}
\newcommand{\bi}{\boldsymbol{i}}\newcommand{\bI}{\boldsymbol{I}}
\newcommand{\bj}{\boldsymbol{j}}\newcommand{\bJ}{\boldsymbol{J}}
\newcommand{\bk}{\boldsymbol{k}}\newcommand{\bK}{\boldsymbol{K}}
\newcommand{\bl}{\boldsymbol{l}}\newcommand{\bL}{\boldsymbol{L}}
\newcommand{\bm}{\boldsymbol{m}}\newcommand{\bM}{\boldsymbol{M}}
\newcommand{\bn}{\boldsymbol{n}}\newcommand{\bN}{\boldsymbol{N}}
\newcommand{\bo}{\boldsymbol{o}}\newcommand{\bO}{\boldsymbol{O}}
\newcommand{\bp}{\boldsymbol{p}}\newcommand{\bP}{\boldsymbol{P}}
\newcommand{\bq}{\boldsymbol{q}}\newcommand{\bQ}{\boldsymbol{Q}}
\newcommand{\br}{\boldsymbol{r}}\newcommand{\bR}{\boldsymbol{R}}
\newcommand{\bs}{\boldsymbol{s}}\newcommand{\bS}{\boldsymbol{S}}
\newcommand{\bt}{\boldsymbol{t}}\newcommand{\bT}{\boldsymbol{T}}
\newcommand{\bu}{\boldsymbol{u}}\newcommand{\bU}{\boldsymbol{U}}
\newcommand{\bv}{\boldsymbol{v}}\newcommand{\bV}{\boldsymbol{V}}
\newcommand{\bw}{\boldsymbol{w}}\newcommand{\bW}{\boldsymbol{W}}
\newcommand{\bx}{\boldsymbol{x}}\newcommand{\bX}{\boldsymbol{X}}
\newcommand{\by}{\boldsymbol{y}}\newcommand{\bY}{\boldsymbol{Y}}
\newcommand{\bz}{\boldsymbol{z}}\newcommand{\bZ}{\boldsymbol{Z}}

\newcommand{\balpha}{\boldsymbol{\alpha}}\newcommand{\bAlpha}{\boldsymbol{\Alpha}}
\newcommand{\bbeta}{\boldsymbol{\beta}}\newcommand{\bBeta}{\boldsymbol{\Beta}}
\newcommand{\bgamma}{\boldsymbol{\gamma}}\newcommand{\bGamma}{\boldsymbol{\Gamma}}
\newcommand{\bdelta}{\boldsymbol{\delta}}\newcommand{\bDelta}{\boldsymbol{\Delta}}
\newcommand{\bepsilon}{\boldsymbol{\epsilon}}\newcommand{\bEpsilon}{\boldsymbol{\Epsilon}}
\newcommand{\bzeta}{\boldsymbol{\zeta}}\newcommand{\bZeta}{\boldsymbol{\Zeta}}
\newcommand{\beeta}{\boldsymbol{\eta}}\newcommand{\bEta}{\boldsymbol{\Eta}} %
\newcommand{\btheta}{\boldsymbol{\theta}}\newcommand{\bTheta}{\boldsymbol{\Theta}}
\newcommand{\biota}{\boldsymbol{\iota}}\newcommand{\bIota}{\boldsymbol{\Iota}}
\newcommand{\bkappa}{\boldsymbol{\kappa}}\newcommand{\bKappa}{\boldsymbol{\Kappa}}
\newcommand{\blambda}{\boldsymbol{\lambda}}\newcommand{\bLambda}{\boldsymbol{\Lambda}}
\newcommand{\bmu}{\boldsymbol{\mu}}\newcommand{\bMu}{\boldsymbol{\Mu}}
\newcommand{\bnu}{\boldsymbol{\nu}}\newcommand{\bNu}{\boldsymbol{\Nu}}
\newcommand{\bxi}{\boldsymbol{\xi}}\newcommand{\bXi}{\boldsymbol{\Xi}}
\newcommand{\bomikron}{\boldsymbol{\omikron}}\newcommand{\bOmikron}{\boldsymbol{\Omikron}}
\newcommand{\bpi}{\boldsymbol{\pi}}\newcommand{\bPi}{\boldsymbol{\Pi}}
\newcommand{\brho}{\boldsymbol{\rho}}\newcommand{\bRho}{\boldsymbol{\Rho}}
\newcommand{\bsigma}{\boldsymbol{\sigma}}\newcommand{\bSigma}{\boldsymbol{\Sigma}}
\newcommand{\btau}{\boldsymbol{\tau}}\newcommand{\bTau}{\boldsymbol{\Tau}}
\newcommand{\bypsilon}{\boldsymbol{\ypsilon}}\newcommand{\bYpsilon}{\boldsymbol{\Ypsilon}}
\newcommand{\bphi}{\boldsymbol{\phi}}\newcommand{\bPhi}{\boldsymbol{\Phi}}
\newcommand{\bchi}{\boldsymbol{\chi}}\newcommand{\bChi}{\boldsymbol{\Chi}}
\newcommand{\bpsi}{\boldsymbol{\psi}}\newcommand{\bPsi}{\boldsymbol{\Psi}}
\newcommand{\bomega}{\boldsymbol{\omega}}\newcommand{\bOmega}{\boldsymbol{\Omega}}

\newcommand{\nA}{\mathbb{A}}
\newcommand{\nB}{\mathbb{B}}
\newcommand{\nC}{\mathbb{C}}
\newcommand{\nD}{\mathbb{D}}
\newcommand{\nE}{\mathbb{E}}
\newcommand{\nF}{\mathbb{F}}
\newcommand{\nG}{\mathbb{G}}
\newcommand{\nH}{\mathbb{H}}
\newcommand{\nI}{\mathbb{I}}
\newcommand{\nJ}{\mathbb{J}}
\newcommand{\nK}{\mathbb{K}}
\newcommand{\nL}{\mathbb{L}}
\newcommand{\nM}{\mathbb{M}}
\newcommand{\nN}{\mathbb{N}}
\newcommand{\nO}{\mathbb{O}}
\newcommand{\nP}{\mathbb{P}}
\newcommand{\nQ}{\mathbb{Q}}
\newcommand{\nR}{\mathbb{R}}
\newcommand{\nS}{\mathbb{S}}
\newcommand{\nT}{\mathbb{T}}
\newcommand{\nU}{\mathbb{U}}
\newcommand{\nV}{\mathbb{V}}
\newcommand{\nW}{\mathbb{W}}
\newcommand{\nX}{\mathbb{X}}
\newcommand{\nY}{\mathbb{Y}}
\newcommand{\nZ}{\mathbb{Z}}

\newcommand{\cA}{\mathcal{A}}
\newcommand{\cB}{\mathcal{B}}
\newcommand{\cC}{\mathcal{C}}
\newcommand{\cD}{\mathcal{D}}
\newcommand{\cE}{\mathcal{E}}
\newcommand{\cF}{\mathcal{F}}
\newcommand{\cG}{\mathcal{G}}
\newcommand{\cH}{\mathcal{H}}
\newcommand{\cI}{\mathcal{I}}
\newcommand{\cJ}{\mathcal{J}}
\newcommand{\cK}{\mathcal{K}}
\newcommand{\cL}{\mathcal{L}}
\newcommand{\cM}{\mathcal{M}}
\newcommand{\cN}{\mathcal{N}}
\newcommand{\cO}{\mathcal{O}}
\newcommand{\cP}{\mathcal{P}}
\newcommand{\cQ}{\mathcal{Q}}
\newcommand{\cR}{\mathcal{R}}
\newcommand{\cS}{\mathcal{S}}
\newcommand{\cT}{\mathcal{T}}
\newcommand{\cU}{\mathcal{U}}
\newcommand{\cV}{\mathcal{V}}
\newcommand{\cW}{\mathcal{W}}
\newcommand{\cX}{\mathcal{X}}
\newcommand{\cY}{\mathcal{Y}}
\newcommand{\cZ}{\mathcal{Z}}

\newcommand{\figref}[1]{Fig.~\ref{#1}}
\newcommand{\secref}[1]{Section~\ref{#1}}
\newcommand{\algref}[1]{Algorithm~\ref{#1}}
\newcommand{\eqnref}[1]{Eq.~\eqref{#1}}
\newcommand{\tabref}[1]{Table~\ref{#1}}

\def\mc{\mathcal}
\def\mb{\boldsymbol}

\newcommand{\T}{^{\raisemath{-1pt}{\mathsf{T}}}}

\newcommand{\Perp}{\perp\!\!\! \perp}

\makeatletter
\DeclareRobustCommand\onedot{\futurelet\@let@token\@onedot}
\def\@onedot{\ifx\@let@token.\else.\null\fi\xspace}
\def\eg{e.g\onedot,\xspace} \def\Eg{E.g\onedot,\xspace}
\def\ie{i.e\onedot,\xspace} \def\Ie{I.e\onedot,\xspace}
\def\cf{cf\onedot} \def\Cf{Cf\onedot}
\def\etc{etc\onedot}
\def\vs{vs\onedot}
\def\wrt{wrt\onedot}
\def\dof{d.o.f\onedot}
\def\etal{et~al\onedot}
\def\iid{i.i.d\onedot}
\def\evs{\emph{vs}\onedot}
\makeatother

\newcommand*\rot{\rotatebox{90}}

\newcommand{\boldparagraph}[1]{\vspace{0.4em}\noindent{\bf #1:}}

\definecolor{darkgreen}{rgb}{0,0.7,0}
\definecolor{lightred}{rgb}{1.,0.5,0.5}

\crefname{section}{Sec.}{Secs.}
\Crefname{section}{Section}{Sections}
\Crefname{table}{Table}{Tables}
\crefname{table}{Tab.}{Tabs.}

\twocolumn[\maketitle%

\vspace{-1cm}

\begin{center}
    \includegraphics[trim={0px 0px 0px 0px}, clip, width=\linewidth]{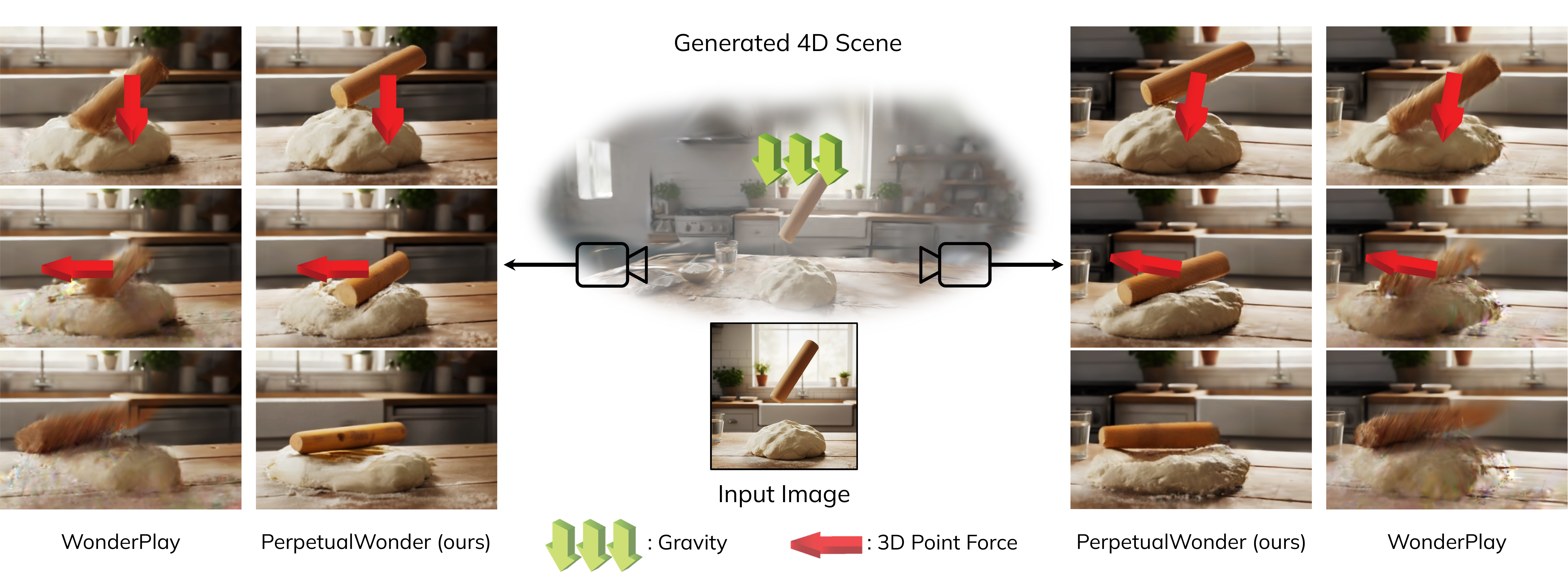}
\end{center}
\vspace{-0.7cm}
\captionof{figure}{
We propose \textbf{\ourmethod}, a hybrid generative simulator that generates a 4D scene with long-horizon actions and a single image. 
Here we show a side-by-side comparison for a three-step action sequence~(top to bottom, actions overlaid on the images). The left and right image blocks show renderings from two different viewpoints.
\ourmethod shows superior performance over the previous method.
We show video results in \url{https://johnzhan2023.github.io/PerpetualWonder/}.
}%
\label{fig:teaser}\bigbreak]

{
\let\thefootnote\relax\footnotetext{$^*$Equal contribution. $^{\text{†}}$Work was done when J. Zhan was a visiting student at Stanford University. J. Zhan is currently with Fudan University.}
}

\begin{abstract}

\vspace{-0.5em}

We introduce \ourmethod, a hybrid generative simulator that enables long-horizon, action-conditioned 4D scene generation from a single image. Current works fail at this task because their physical state is decoupled from their visual representation, which prevents generative refinements to update the underlying physics for subsequent interactions. \ourmethod solves this by introducing the first true closed-loop system. It features a novel unified representation that creates a bidirectional link between the physical state and visual primitives, allowing generative refinements to correct both the dynamics and appearance. It also introduces a robust update mechanism that gathers supervision from multiple viewpoints to resolve optimization ambiguity. Experiments demonstrate that from a single image, \ourmethod can successfully simulate complex, multi-step interactions from long-horizon actions, maintaining physical plausibility and visual consistency.

\end{abstract}    
\vspace*{-8pt}\section{Introduction}
\label{sec:intro}

Recent years have seen remarkable progress in generative models for text, images, and videos. This rapid advancement motivates the creation of generative world models~\cite{sora,hunyuanvoyager,hunyuanworld}, which are crucial for applications in VR/AR, gaming, and embodied AI~\cite{qureshi2025splatsim}. A key capability for such models is not just to generate realistic content, but to simulate a world that responds to user actions. We study the task of \textbf{action-conditioned 4D scene generation from a single image}. Given a single input image and a sequence of physical actions~(local forces like pushes and pokes, or global forces like wind fields and gravity), the goal is to generate the dynamic 4D scene that corresponds to the actions and evolves plausibly over time.

Early attempts to generate 4D content in response to actions relied heavily on traditional physical simulation~\cite{physgaussian,physdreamer,physgen3d,dreamphysics}. These methods, while offering precise and interpretable physical control, are driven entirely by the traditional simulator for both dynamics and appearance. This often results in a significant realism gap, as simplified physics and analytic rendering struggle to capture the complex visual phenomena of the real world, such as subtle material deformations, lighting changes, and secondary visual effects like splashes. Concurrently, modern video generation models~\cite{sora,veo3} have become incredibly powerful, learning strong priors about real-world dynamics and appearance from massive video data. 

This presents a new opportunity, leading to the rise of the \textbf{hybrid generative simulator}~\cite{Wonderplay,physmotion}: a system that first uses the traditional physical simulation to generate coarse, action-conditioned dynamics, and then employs a video generation model as a neural refiner to achieve high-fidelity visual realism. The hybrid generative simulator aims for the best of both worlds, combining the strengths of traditional physical simulators, including consistency and controllability, with the power of video generation models, which provide visual realism and complex dynamics. 

Recent WonderPlay~\cite{Wonderplay} is a realization of this hybrid generative simulator concept.
However, its approach is fundamentally limited to short-term interactions within a single time window. The core problem is that the flow of information is incomplete: the physical state informs the video model, but the video model's refinement only propagates back to the scene's appearance representation, not its underlying physical state. 
The physical and visual representations are thus decoupled. This prevents any form of long-horizon, sequential interaction, as the physical simulator is blind to the generative corrections from the previous step, leading to the accumulation of errors.

We aim to overcome this fundamental limitation and enable long-horizon, sequential actions. This requires a system that can perpetually cycle between user actions, physical simulation, and generative refinement. We identify two fundamental challenges: the first is that current physical states~(physics particle position and velocity) cannot be updated by the refinement from the video generation model. A new representation is required to unify the physical and visual domains. Then, to update the unified representation, the refinement from video generation models must be multi-view to prevent ambiguity in optimization. However, the video models naturally will not generate perfectly consistent videos from different viewpoints. To resolve this ambiguity, a robust update mechanism is required.

To address these challenges, we propose \textbf{\ourmethod}, a new hybrid generative simulator for long-horizon action-conditioned 4D scene generation, as shown in Figure~\ref{fig:teaser}. First, we introduce \textit{visual-physical aligned particle}~(VPP), a novel unified representation that tightly binds physics particles to the visual representations. The proposed VPP acts as a bidirectional bridge: enabling the forward physics pass that uses physical simulation to drive the visual prediction, and critically, updating the physics particles through the optimized visual representation in a backward optimization, resulting in an innovative closed-loop system. Next, we propose a multi-view optimization mechanism to ensure the update is 3D consistent and plausible. We first initialize a complete 3D scene from the input image using dense view generation. This initialization allows us to render the scene from arbitrary viewpoints in a wide range and use the video model to gather supervision from multiple views. Then we progressively leverage refined videos from multiple viewpoints for backward optimization. This strategy resolves ambiguity, producing a 4D scene that is both visually realistic and physically coherent, ready for the next user action. In summary, our contributions are:

\begin{itemize}
    \item We tackle the task of long-horizon action-conditioned 4D scene generation, enabling sequential action interactions.
    \item We propose \ourmethod, a novel hybrid generative simulator that features a unified representation for both physical state and visual appearance, and a multi-view optimization mechanism for consistent scene updates.
    \item We demonstrate that \ourmethod consistently outperforms prior work in action-conditioned 4D scene generation, including both long-horizon interaction abilities and scene consistency.
\end{itemize}

\begin{figure*}
  \centering
  \includegraphics[width=\linewidth]{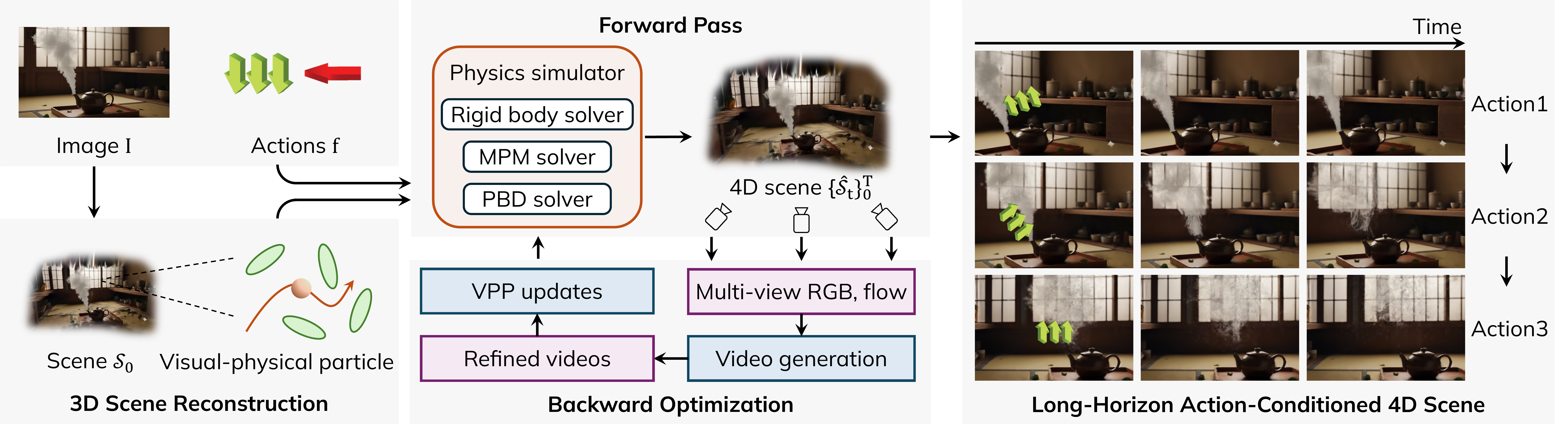}
  \vspace{-2em}
  \caption{\textbf{Overview of \ourmethod}. Given an input image, based on the visual-physical aligned particle, we reconstruct a 3D scene from synthesized dense views. Then we iterate between a forward physics pass and a backward neural optimization. The forward pass leverages physical simulation to generate coarse scene dynamics. Then the backward optimization updates the scene according to the multi-view refined videos from the video generation model. The closed-loop system enables long-horizon actions for the final 4D scene generation. The rendered results on the right showcase the generated scene from each consecutive action.}
  \label{fig:pipeline}
  \vspace{-1.2em}
\end{figure*}
\section{Related Work}
\label{sec:related_work}

\myparagraph{Dynamic 4D scene generation.}
Our work generates action-conditioned 4D scenes and connects to a rich body of research on dynamic scene representation and generation. Early work in this domain focused on reconstructing dynamic scenes from real-world captures. Representations rapidly evolved from dynamic Neural Radiance Fields~(NeRF)~\cite{nerf,pumarola2021d,park2021hypernerf,park2021nerfies} to dynamic Gaussian Splatting~\cite{3dgs,4dgs_1,4dgs_2,4dgs_3}. While these methods achieve high-fidelity rendering of complex motion, they are fundamentally limited to replaying pre-captured events and do not support user actions or the simulation of novel dynamics.

More recently, a dominant line of research has focused on generative models for synthesizing novel 4D content. Many of these approaches distill the powerful priors from large-scale video models to generate 4D animations from text or image prompts~\cite{tc4d,4dfy,alignyourgaussians,dreamgaussian4d,textto4d,stag4d,geng2025birth}. These methods leverage dynamic 3D representations to create temporally consistent animations. Other works focus on directly modeling the 4D space-time volume~\cite{controllingspaceandtime,wu2024cat4d,genxd} or parameterizing 4D representations with generative networks~\cite{l4gm,gaussianvae}. However these approaches share a critical limitation: the synthesized dynamics are passive. They generate pre-determined animations and lack the mechanisms to simulate diverse, physically plausible responses to user input actions, which is the core focus of our work. 

\myparagraph{Physics-grounded scene generation.}
To enable action-conditioned interactions, a separate line of work has focused on integrating physical principles~\cite{le2025pixie} and traditional physical simulation methods into the scene generation process. Early methods relied entirely on traditional physical simulation~\cite{physgen3d,dreamphysics,physgaussian,physdreamer}, which provides precise, interpretable control but suffers from a significant realism gap. These simulators often use simplified, approximated physics and rendering with fixed visual appearance, struggling to capture the visual phenomena of the real world, such as subtle material deformations and secondary visual effects.

To bridge this realism gap, recent works have begun to integrate physics with the strong priors of generative models. This has culminated in the hybrid generative simulator~\cite{Wonderplay,physmotion} approaches, which aim to blend the controllability of the physical simulators with the visual realism of the video generation models. The most relevant work, WonderPlay~\cite{Wonderplay}, first uses the physics solvers to generate coarse, action-conditioned dynamics and then employs a video model as a neural refiner to achieve high-fidelity visual realism.
However, these methods are limited by a fundamental architectural drawback that the flow of information is incomplete. The generative refinements, whether for appearance or dynamics, only affect the visual primitives and do not propagate back to the underlying physical state. Furthermore, this makes these methods fundamentally limited to short-term interactions, and are unable to support the long-horizon actions that our work aims to solve.

\myparagraph{Controllable video generation.}
Concurrently, video generation models have become incredibly powerful~\cite{sora,stepvideo,Wan,veo3,Cogvideo,opensora_1}, achieving stunning realism, but controlling their output remains a significant challenge. 
Most existing control methods focus on non-physical aspects, such as following text instructions, camera trajectories~\cite{recammaster,Gen3c,cinemaster,stablevirtualvideo}, or 2D motion guidance through keypoints and trajectories~\cite{Motion_Conditioned,motionprompting,mofa,motioni2v,draganything,tora}. While some works explore conditioning on 2D force vectors~\cite{Force_prompt} to mimic the real-world actions, these models lack an explicit underlying 3D representation. These 2D-centric video approaches are insufficient for our task, as they cannot ensure physically accurate action conditions within a 3D scene or guarantee 3D consistency when rendering the resulting 4D scenes from novel viewpoints. In contrast, our method operates on a full 3D representation, providing the structure for both precise physical interaction and consistent multi-view rendering.

\section{PerpetualWonder}
\label{s:method}

Our goal is to achieve long-horizon action-conditioned 4D scene generation from a single image $\mathbf{I}$. Given a sequence of user actions $\{\mathcal{A}_t\}_{t=0}^{T-1}$ including the global force $\mathbf{f}(x,y,z,t)$~(\eg gravity, wind field) and/or the local force $\mathbf{f}(t)$, our proposed \ourmethod outputs a dynamic 4D scene sequence $\{\mathcal{S}_t\}_{t=0}^T$. At any time $t$, the scene state $\mathcal{S}_{t}=(\mathcal{B}_{t},\mathcal{F}_{t})$ is decomposed into the background $\mathcal{B}_{t}$ and the dynamic, interactable foreground $\mathcal{F}_{t}$.

As illustrated in Figure~\ref{fig:pipeline}, \ourmethod achieves this by an innovative closed-loop hybrid generative simulator, perpetually iterating between a forward physics pass $\Phi_{p}$ and a backward neural optimization pass $\Psi_{n}$. To enable this, we must solve the two fundamental challenges: first, crafting a unified representation that allows the physical state to be updated by generative refinements from video models; and second, developing a robust and consistent optimization mechanism to perform these updates without ambiguity.

Accordingly, we first introduce the proposed \textit{visual-physical aligned particle}~(VPP) as a unified representation~(Section~\ref{s:method-vpp}). We then detail our multi-view optimization mechanism~\ref{s:method-updates}, which leverages a single image-based dense 3D reconstruction to perform a progressive, multi-view backward optimization. Finally, we assemble these components into the complete \ourmethod simulation loop for long-horizon actions~(Section~\ref{s:method-loop}).

\subsection{Visual-Physical Aligned Particle}
\label{s:method-vpp}

The core of long-horizon action interactions lies in the underlying representation. Previous hybrid generative simulators use decoupled visual primitives~(\eg gaussian splatting~\cite{3dgs}) for appearance and physics particles for dynamics. The incomplete binding uses physics particles to drive visual primitives, making it impossible for visual refinements from the video generation model to correct the underlying dynamics. This prevents a closed-loop system and makes long-horizon simulation inapplicable.

To solve this, we introduce the VPP, a novel unified representation that tightly binds the physics particles to the visual primitives, creating a bidirectional bridge between dynamics and appearance.

We define the foreground $\mathcal{F}_{t}$ as a set of $O$ objects, $\mathcal{F}_{t}=\{\mathcal{P}_{t}^{o},\mathcal{V}_{t}^{o},\mathcal{G}_{t}^{o}\}_{o=1}^{O}$. For simplicity, here we omit the object index $o$ and time index $t$. The VPP for a single object consists of:

\vspace{0.1cm}
\noindent
\underline{Physics dynamics.} 
A set of $J$ physics particle positions $\mathcal{P}=\{p_{j}\}_{j=1}^{J}$ and their velocities $\mathcal{V}=\{v_{j}\}_{j=1}^{J}$. The particles $\mathcal{P}$ are sampled from an initial object mesh volume. We detail the process for obtaining the object mesh in the following 3D scene initialization subsection.

\vspace{0.1cm}
\noindent
\underline{Visual appearance.} 
A set of $J \times K$ visual primitives, \ie gaussians~\cite{3dgs}, $\mathcal{G}=\bigcup_{j=1}^{J}\{g_{j,k}\}_{k=1}^{K}$. Specifically, each physics particle $p_j$ serves as an anchor to a small set of $K$ gaussian primitives.

The key to the VPP is how these $K$ gaussians are parameterized relative to their anchor particle $p_j$, enabling both adherence to physics and optimization of appearance and dynamics. We define the attributes of these gaussians with the following details:

\begin{itemize}
    \item \textbf{Position offset $\tilde{p}$}: Each gaussian's 3D position $\mu_{j,k}$ is defined by a small, learnable position offset $\tilde{p}_{j,k}$ from its corresponding physics particle $p_{j}$:
    \begin{equation}
    \mu_{j,k}=p_{j}+\tanh(\tilde{p}_{j,k})\cdot\delta,
  \label{eq:pos_offset}
    \end{equation}
    where $\delta$ is the physics particle size defined during the simulator's sampling process.

    \item \textbf{Scale}: Each gaussian primitive is defined to be isotropic, with a fixed scaling value not larger than $\delta$.

    \item Spatial opacity $o_{s}$ and temporal opacity $o_{t}(t)$: Inspired by \citep{freetimegs}, we define two opacity parameters. $o_{s}$ is the standard learnable spatial opacity scalar~\cite{3dgs}. $o_{t}(t)$ is a learnable temporal opacity, parameterized by a center time $\mu_{t}$ and duration $s_{d}$:
    \begin{equation}
        o_{t}(t)=\exp\left(-\frac{1}{2}\times\left(\frac{t-\mu_{t}}{s_{d}}\right)^{2}\right).
    \label{eq:temp_opacity}
    \end{equation}
    The final opacity is $o(t)=o_{s}\times o_{t}(t)$.
    
\end{itemize}

Rotation and color attributes are parameterized the same as in the original 3D Gaussian Splatting work~\cite{3dgs}. This VPP representation forms the foundation of our closed-loop system. 
By ensuring every physics particle $p_j$ has a corresponding set of visual primitives $\{g_{j,k}\}$, all dynamics and appearance are now expressed by the optimizable visual primitives. This structure creates the bidirectional bridge, where the forward physics pass drives the dynamics by updating $p_j$, which in turn moves all anchored visual primitives. Critically, the backward optimization pass can now refine the final 4D scene by optimizing the attributes of $\{g_{j,k}\}$  while being constrained to remain consistent with the anchoring physics particles.

\subsection{Multi-View Optimization}
\label{s:method-updates}

The proposed VPP serves as a representation that can be updated for both 4D scene dynamics and appearance. The remaining challenge lies in how to consistently perform this update. Simply optimizing from a single-view video refinement, as done in WonderPlay~\cite{Wonderplay}, leads to significant ambiguity and artifacts from novel viewpoints.
To tackle this, we introduce a robust multi-view optimization mechanism. This mechanism consists of two components. First, we initialize the complete 3D scene from the single input image to enable rendering from arbitrary viewpoints. Second, we leverage this 3D scene to perform a progressive optimization using supervision gathered from multiple views.

\myparagraph{3D scene initialization.}
As the initialization for our hybrid generative simulator, we first reconstruct the 3D scene $\mathcal{S}_{0}$ from the single input image $I$. To construct a 3D scene that supports rendering from arbitrary viewpoints, we employ a state-of-the-art camera-controlled video model GEN3C~\cite{Gen3c} to synthesize dense surrounding views of the scene from the input image. This video is then processed with COLMAP~\cite{sfm} to acquire a scene point cloud for initializing 3D Gaussians.

Following standard 3DGS optimization~\cite{3dgs}, we obtain the set of $N$ gaussian primitives $\{G_{i}\}_{i=1}^{N}$ as the initial scene representation. Each gaussian $G_i$ is parameterized by position $p_{i}$, orientation $q_{i}$, scale $s_{i}$, opacity $o_{i}$, and color $c_{i}$.

To decompose the scene into background and foreground objects, inspired by Gaussian Grouping~\cite{gaussian_grouping}, we add a learnable feature $g_{i}$ to each primitive. We leverage SAM2~\cite{sam2} to obtain object masks on the dense surrounding views, which supervise these learnable features. After decomposition, the Gaussian primitives are split into a background set $\mathcal{B}_{0}$ and sets for each foreground object. These foreground object gaussians are then transformed into closed-surface meshes using TSDFusion~\cite{tsdfusion}. Then these meshes are used to sample the initial physics particles $\mathcal{P}_{0}$ for the proposed VPP. This is followed by another round of optimization for the gaussian primitives $\{\mathcal{G}_0^o\}_{o=1}^O$, with respect to the frames from GEN3C~\cite{Gen3c}, and replace the original foreground objects' gaussians.

This approach contrasts with the scene initialization in WonderPlay, which relies on single-view depth-unprojection~\cite{wonderjourney,wonderworld} and object placement. Our multi-view reconstruction process builds the background and all objects together in a single, unified 3D coordinate space, which is essential for rendering from dense, arbitrary viewpoints.

\myparagraph{Progressive multi-view optimization.}
The refinement process $\Psi_n$ leverages pre-trained video generation models to refine both appearance and dynamics of the underlying 4D scene. The bridge between our VPP representation and the video generation model is the rendered video. Following \citep{Wonderplay}, the coarse 4D scene from physical simulation~(detailed in Section~\ref{s:method-loop}) is rendered into RGB and optical flow frames from a chosen viewpoint. This coarse video is then refined by the video generation model~\cite{Go_with_the_flow,Cogvideo} through a bimodal control scheme, resulting in a refined video $\mathbf{V}_{t}$.

However, refined videos from different viewpoints are inevitably inconsistent and can not be directly used for optimization. To resolve the ambiguity and acquire the desired 4D scene,  we introduce a two-part solution.

First, we design a loss function that, combined with our VPP representation, provides a strong consistency prior. For a given time step $t$, the overall loss function is:

\begin{align}
    \mathcal{L}&=\mathcal{L}_{\mathrm{p}}(\texttt{Render}(\mathcal{B}_{t})\odot(1-\mathbf{M}),\mathbf{V}_{t}\odot(1-\mathbf{M}))\notag \\
&+\mathcal{L}_{\mathrm{p}}(\texttt{Render}(\mathcal{G}_{t}),\mathbf{V}_{t}\odot M)+\lambda_{\mathrm{sim}}\mathcal{L}_{\mathrm{sim}}.
\label{eq:main_loss}
\end{align}

Here, $\mathbf{M}$ is the binary mask for the foreground VPPs, $\texttt{Render}(\cdot)$ is the gaussian rendering function, and $\mathcal{L}_{p}$ is the photometric loss~(L1 and SSIM). In practice, we also model the background gaussians $\mathcal{B}_{t}$ with learnable spatial and temporal opacity (Eq.~\ref{eq:temp_opacity}) and other attributes~(excluding position) to capture secondary visual effects like shadows. For the foreground VPP $\mathcal{G}_{t}$, we introduce a simulation consistency loss term $\mathcal{L}_{\mathrm{sim}}$:
\begin{equation}
    \mathcal{L}_{\mathrm{sim}} = \frac{1}{T \cdot J} \sum_{t=1}^{T} \sum_{j=1}^{J} \left\Vert p_{j,t} - \frac{1}{K}\sum_{k=1}^{K} \mu_{j,k,t} \right\Vert_2^2
\label{eq:sim_loss}
\end{equation}

The $\mathcal{L}_{\mathrm{sim}}$ penalizes the visual primitives $\mu_{j,k,t}$ for deviating from their corresponding physics particle $p_{j,t}$. The VPP representation and the simulation consistency loss act as a strong regularizer, ensuring the optimized visual primitives do not break apart from their physical anchors, which inherently mitigates inconsistency.

Second, to resolve the remaining visual ambiguity from conflicting refinements in a multi-view setting, we introduce a progressive optimization strategy. To leverage inconsistent multi-view videos without introducing artifacts, we first render and refine the video only from the input image's viewpoint and optimize the scene representation with respect to this single video. Then, we render the 4D scene from other viewpoints and use the video model to refine them with a smaller control weight. Finally, we leverage all refined videos from every viewpoint to optimize the scene representation again, yielding a consistent 4D scene.

\subsection{Simulation Loop}
\label{s:method-loop}

With the proposed novel VPP representation and multi-view optimization mechanism now defined, we can assemble the  \ourmethod simulation loop. This loop operates over a time window of $T$ steps and consists of three key stages: a forward pass to generate the entire sequence, a backward optimization pass to refine it, and a loop closure that enables the next round of action-conditioned interaction.

\myparagraph{Forward pass.}
The loop begins with the forward physics pass $\Phi_{p}$. Given the scene state $\mathcal{S}_{0}$, the hybrid generative simulator computes the coarse 4D scene for the time window of $T$ steps. This is achieved by applying the physics operator $\Phi_{p}$ step-by-step for each user action $\mathcal{A}_{t}$ from $t=0$ to $T-1$: $\hat{\mathcal{S}}_{t+1}=\Phi_{p}(\hat{\mathcal{S}}_{t},\mathcal{A}_{t})$. This process generates a coarse sequence $\{\hat{\mathcal{S}}_{t}\}_{t=0}^{T}$. We adopt a set of solvers~\cite{Wonderplay,muller2005meshless,jiang2016material,muller2007position} for various materials, including cloth, sand, snow, liquid, smoke, elastic, and rigid bodies. We refer the reader to \citet{Wonderplay} for more details of the physical simulation process. As aforementioned, this forward pass provides controllable dynamics in response to actions but lacks visual realism and may contain physical inaccuracies.

\myparagraph{Backward optimization.}
The coarse sequence $\{\hat{\mathcal{S}}_{t}\}_{t=0}^{T}$ is then fed into the refinement process $\Psi_n$. As introduced in Section~\ref{s:method-updates}, we apply our progressive multi-view optimization to the coarse 4D scene. This step leverages video model priors to correct the appearance and dynamics across all $T$ steps, resulting in the final refined sequence $\{\mathcal{S}_{t}\}_{t=0}^{T}$.

\begin{figure*}
  \centering
  \includegraphics[width=\linewidth]{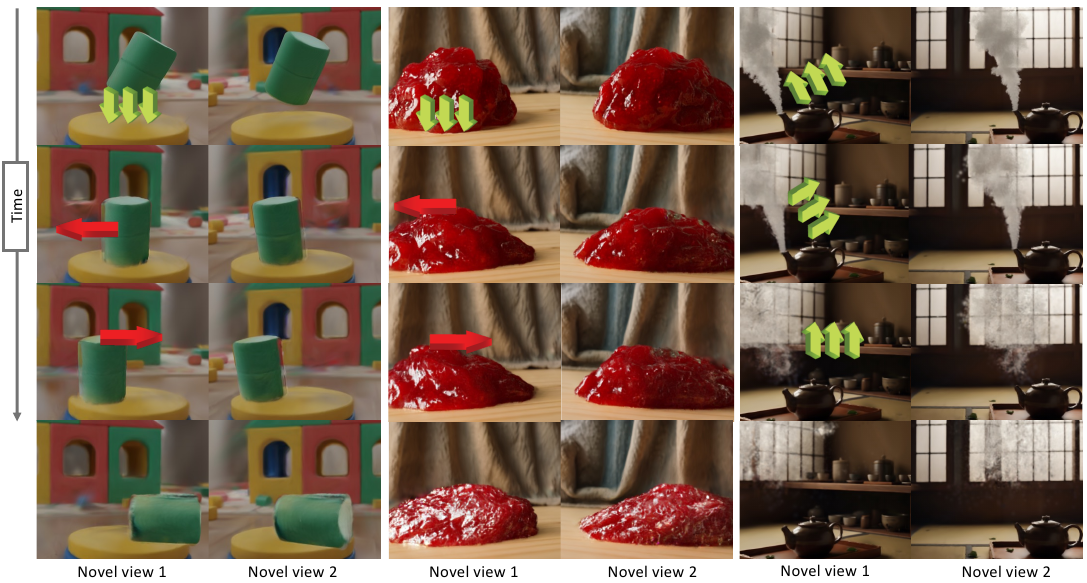}
   \vspace{-2em}
  \caption{\textbf{Qualitative results} of the proposed \ourmethod. We show the long-horizon scenes with three consecutive actions. \myicongravity{} and \myiconforce indicate global force~(gravity or wind force field), and 3D point force, respectively. The results are all rendered from novel views, demonstrating our method's ability in long-horizon action-conditioned 4D scene generation.}
  \label{fig:quality_result}
  \vspace{-1.5em}
\end{figure*}

\myparagraph{Loop closure and long-horizon actions.}
The final step closes the loop and enables perpetual interaction, \ie performing a new round of actions for the following time window. The final refined state of the current time window, $\mathcal{S}_{T}$, becomes the starting state $\mathcal{S}_{0}$ for the next $T$-step simulation. To prepare the system for this new time window, we update the VPP's underlying physics particles using the results of the backward pass. Specifically, we average the positions of the optimized visual primitives $\{g_{j,k}\}$ at time $T$ to update the position $p_j$ of their corresponding physics particle. The velocity is directly inherited from the original velocities at time $T$, which is applicable because $\mathcal{L}_{\mathrm{sim}}$ limits the position updates of physics particles in a small range.  This corrected physics state $\{\mathcal{P}_{T},\mathcal{V}_{T}\}$ becomes the input for the next forward pass sequence, allowing the system to simulate sequential interactions over a long horizon.
\section{Experiment}

\begin{table}[t]
  \centering
  \small
    \setlength{\tabcolsep}{4pt}
  \begin{tabular}{l ccc}
    \toprule
Method& Camera Ctrl & 3D Consist& Imaging\\
    \midrule
    Wan2.2~\cite{Wan}& 59.73& 65.35& \underline{67.03}\\
    GEN3C~\cite{Gen3c}& \underline{80.29}& 61.69& 66.25\\
    WonderPlay~\cite{Wonderplay}& 75.95& 63.93& 36.80\\
        Tora~\cite{tora}& 51.80& 60.77& 54.37\\
    Wan2.6~\cite{Wan}& 64.75& 70.49& 66.09\\
    DaS~\cite{gu2025diffusion}& 78.96& 62.18& 60.23\\
 Veo3.1~\cite{veo3}& 60.61& \underline{73.93}&\textbf{67.82}\\
    \ourmethod (ours) & \textbf{93.26}& \textbf{80.41}& 66.98\\
        
    \bottomrule
  \end{tabular}
  \vspace{-1em}
  \caption{\textbf{Quantitative comparison} on $10$ scenes using WorldScore~\cite{worldscore} metrics. Ctrl~$=$~Controllability, Consist~$=$~Consistency.}
  \label{tab:quantitative} 
  \vspace{-1.5em}
\end{table}

\myparagraph{Implementation details.} 
We initialize the 3D scene $\mathcal{S}_{0}$ by reconstructing it from $242$ views generated by the camera-controlled video model GEN3C~\cite{Gen3c}. We employ Genesis~\cite{Genesis} as our physics simulator for different materials. Our experimental results typically demonstrate multi-step interactions across $3$ time windows, with each window spanning 392 physical simulation steps and receiving different input actions. For refinement, the video generation model~\cite{Go_with_the_flow} is conditioned on RGB and optical flow renderings at the resolution of $H{=}704, W{=}1280$ from the coarse dynamics. The output videos consist of $49$ frames, each frame is sampled from every $8$ physical simulation steps. The progressive multi-view optimization mechanism uses $3$ key views for supervision: the frontal, left-side, and right-side views.

\myparagraph{Baselines.}
We compare \ourmethod against two main categories of state-of-the-art methods: conditional video generation models and hybrid generative simulators. For conditional video generators, we evaluate against Wan2.2~\cite{Wan}, Wan2.6, Veo3.1~\cite{veo3}, Tora~\cite{tora}, DaS~\cite{gu2025diffusion} and GEN3C~\cite{Gen3c}. For I2V video generators, both the action and camera trajectory are specified via text prompts. For GEN3C, we leverage its native camera control capabilities and embed the desired action within the text prompt. For trajectory-based video models like Tora, the dynamics from physical solvers are utilized to drive the generation.
For the hybrid generative simulator, we compare against the most relevant prior work, WonderPlay~\cite{Wonderplay}. To ensure a fair comparison focused on the core simulation loop and representation, we also create a stronger baseline, WonderPlay$++$. This baseline uses our superior multi-view 3D reconstruction for initialization, which provides a more 3D-consistent scene but retains WonderPlay's original decoupled representation and single-view optimization methodology.

\begin{table}[t]
\vspace{.5em}
\centering
\small
\setlength{\tabcolsep}{2pt}
\begin{tabular}{lcc}
\toprule
& Physics Plausibility & Motion Fidelity \\
\midrule
over Wan2.2~\cite{Wan}& 74.1\%& 71.8\%\\
over GEN3C~\cite{Gen3c} & 93.5\%& 83.5\%\\
over WonderPlay~\cite{Wonderplay}& 80.8\%& 86.3\%\\
over Veo3.1~\cite{veo3}& 62.0\%& 70.8\%\\
over Wan2.6~\cite{Wan} & 68.5\%& 77.3\%\\
over Tora~\cite{tora}& 83.5\%& 85.3\%\\
over DaS~\cite{gu2025diffusion}& 80.9\%& 81.9\%\\
\bottomrule
\end{tabular}
\vspace{-.7em}
\caption{\textbf{2AFC human study results} of favor rate of our \ourmethod over baseline methods in dynamic realism.}
\label{tab:human_study}
\vspace{-2em}
\end{table}

\begin{figure*}
  \centering
    \includegraphics[width=\linewidth]{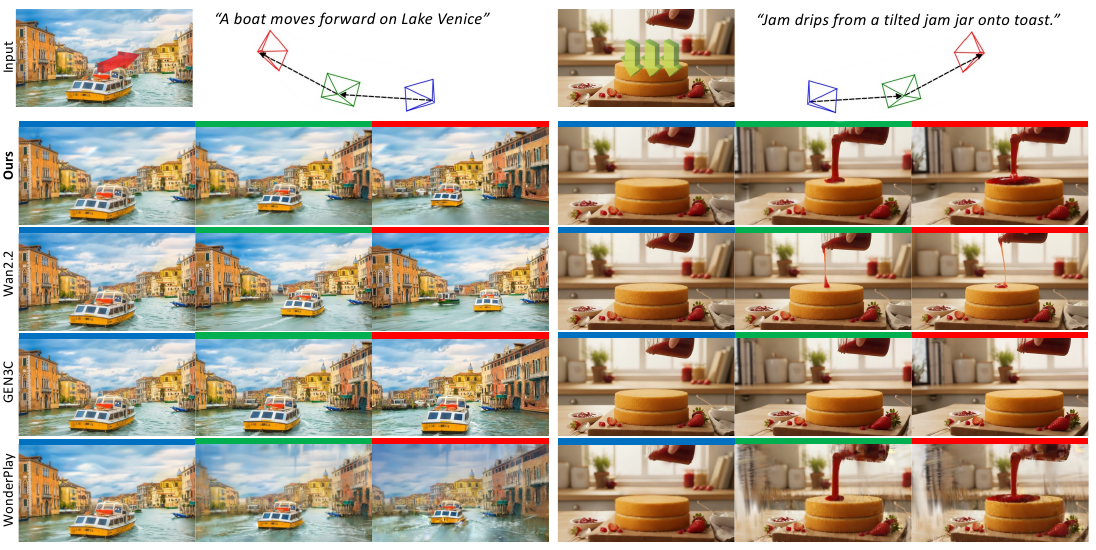}
  \vspace{-2em}
  \caption{\textbf{Qualitative comparisons} between \ourmethod (ours) and the baseline methods. The top row shows the input images, actions, camera trajectories, and the texts describing the actions for conditional video generators~\cite{Wan, Gen3c}. For ease of comparison, only one time window is shown. The images from left to right illustrate the resulting scene dynamics 
and camera motion for each method.}
  \label{fig:baseline}
  \vspace{-1.5em}
\end{figure*}

\myparagraph{Metrics.}
Our evaluation is performed on a curated dataset of $10$ scenes that feature a diverse range of materials, including cloth, rigid bodies, elastic objects, liquids, gases, and granular substances. We assess two primary aspects: the quality of the generated 4D scene and the plausibility of the physical dynamics. 
Please refer to the supplement for the detailed metrics explanations.

\subsection{Results}
\begin{figure*}
  \centering
    \includegraphics[width=\linewidth]{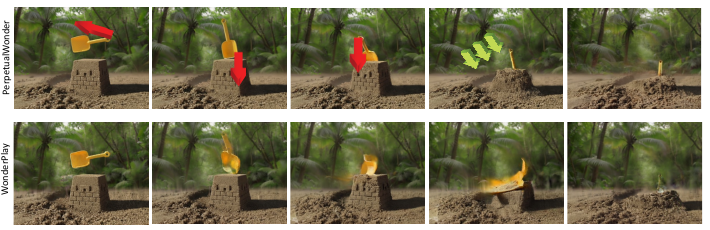}
  \vspace{-2em}
  \caption{\textbf{Long-horizon actions} comparison between \ourmethod~(top row) and WonderPlay~(bottom row). For each method, the view changes across time, illustrating the four-round interaction results on a castle scene. The applied actions are overlaid on the top row.}
  \label{fig:long-horizon}
  \vspace{-1.5em}
\end{figure*}

\begin{figure}
  \centering
  \includegraphics[width=\columnwidth]{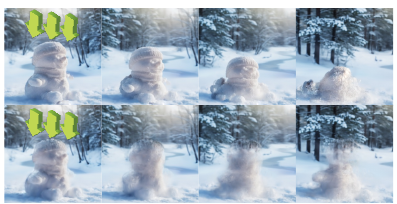}
  \vspace{-2em}
  \caption{\textbf{Ablation} on VPP representation. Top row is with our proposed VPP for the foreground. Bottom row shows using 3D gaussians from the standard Gaussian Splatting~\cite{3dgs} optimization.}
  \label{fig:ablation-vpp}
  \vspace{-1em}
\end{figure}

\myparagraph{Qualitative comparison to baselines.} 
Figure~\ref{fig:baseline} presents a side-by-side qualitative comparison of our method against some representative baselines. As illustrated, the conditional video generation models fail to support both interaction and arbitrary view changes. Wan2.2~\cite{Wan} generates plausible motion but disregards camera change instructions provided in the text prompt, causing the camera to remain almost fixed in both scenes. Conversely, GEN3C~\cite{Gen3c} is 3D-aware and adheres well to camera trajectories, but the object of interest remains static and completely unaffected by the action described in the text prompt.
The hybrid generative simulator WonderPlay~\cite{Wonderplay} successfully applies the user actions, but its reliance on single-view optimization leads to severe visual artifacts and geometric inconsistencies when the scene is rendered from novel viewpoints. In contrast, \ourmethod generates the complete and consistent 4D scene that both correctly responds to the action inputs and naturally supports rendering from arbitrary viewpoints.

\myparagraph{Quantitative results.}
The quantitative results in Table~\ref{tab:quantitative} show \ourmethod's significant advantages over both conditional video generators and the hybrid simulator. \ourmethod achieves best performance in camera controllability and 3D consistency, while keeping a high level of imaging quality.
The user study result is shown in Table~\ref{tab:human_study}. In comparison to all baselines, about 70\% to 90\% of the participants prefer \ourmethod across both aspects. This strong preference provides strong evidence that \ourmethod successfully generates plausible physical dynamics, enabling long-horizon simulations by preventing the accumulated errors that degrade realism in baseline methods.

\myparagraph{Long-horizon actions.} 
A significant advantage of \ourmethod is its ability to handle sequential, long-horizon action inputs within a single scene. We compare with WonderPlay~\cite{Wonderplay} and demonstrate this capability in Figure~\ref{fig:long-horizon} and also Figure~\ref{fig:teaser}.
WonderPlay exhibits severe accumulated errors as the interactions progress. For instance, after the shovel is rotated in the air, it fails to maintain its shape integrity and becomes unrealistically deformed upon insertion into the castle. This is a fundamental limitation of its design: the generative refinements from the last time window do not propagate back to update the physical particles in simulators for the next round of interaction. As a result, at the initialization of each new round, the gaussian primitives are reset to their original positions rather than the optimized positions according to the refined video. The discrepancy causes severe fracturing artifacts and disrupts the temporal continuity across time windows. The decoupling, combined with the reliance on single-view refinement, results in visual artifacts and unstable dynamics as errors compound. 

In contrast, our VPP representation provides a bidirectional bridge, allowing the refined state $\mathcal{S}_{T}$ from the end of one round to become the corrected initial state $\mathcal{S}_{0}$ for the next round. 
As demonstrated in Figure~\ref{fig:quality_result}, which provides more interactive 4D scenes, each with three rounds of interaction, \ourmethod can support sequential interactions for diverse object types such as elastic bodies, gases, and rigid objects, all while generating realistic physical motion.

\subsection{Ablation Study}
We perform an ablation study to assess the respective roles of our VPP representation and the progressive multi-view optimization strategy in generating plausible dynamics and maintaining multi-view consistency.

\begin{figure}
  \centering
  \includegraphics[width=\columnwidth]{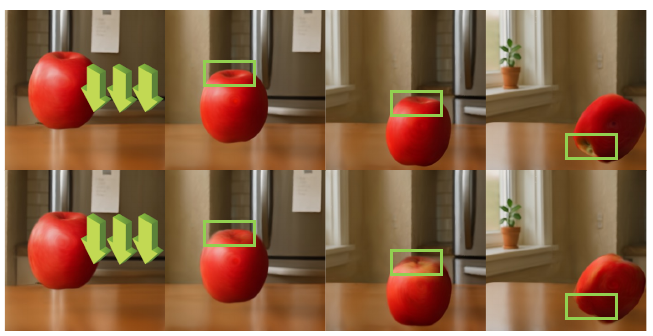}
  \vspace{-2em}
  \caption{\textbf{Ablation} on progressive multi-view optimization. Top row shows the optimized scene using progressive optimization and the bottom row shows direct optimization results.}
  \label{fig:ablation-progressive}
  \vspace{-1.5em}
\end{figure}

\myparagraph{Inherent consistency and plausible dynamics from VPP.}
In the top row of Figure~\ref{fig:ablation-vpp}, we show the dynamic scene represented by VPP after multi-view optimization. The VPP representation constrains visual primitives $\{g_{j,k}\}$ to remain near their corresponding physics particle $p_{j}$. This binding mechanism accurately drives the visual primitives according to the dynamics from physical solvers. We contrast this with a variant~(bottom row) that uses standard 3D gaussian primitives. When subjected to the same multi-view optimization, these unconstrained primitives just aim for minimizing the photometric loss, resulting in degenerate results with chaotic dynamics and visual artifacts. 

\myparagraph{Progressive optimization enhances the multi-view consistency.}
In Figure~\ref{fig:ablation-progressive}, we ablate our progressive optimization strategy~(top row) against a direct optimization variant~(bottom row). In direct optimization, video generators will not naturally generate perfectly consistent multi-view videos. For example, the generated video from the frontal view might hallucinate incorrect visual details (\eg color or shape artifacts), while another view does not, leading to optimization conflicts. By using these inconsistent supervision signals to optimize the underlying scene at once, the representation becomes corrupted, which manifests as blurry textures and appearance flickering on the apple as time progresses. Our progressive approach successfully resolves this ambiguity, yielding a consistent 4D scene.

\section{Conclusion}

We introduce \ourmethod, a novel framework for long-horizon action-conditioned 4D scene generation from a single image. 
\ourmethod enables sequential interactions by unifying the physical and visual representations, and leveraging multi-view optimization for consistent updates.
We demonstrate the superior performance of \ourmethod with diverse scenes and long-horizon actions.

\myparagraph{Acknowledgments.} This work is in part supported by NSF RI \#2211258 and \#2338203, ONR YIP N00014-24-1-2117, ONR MURI N00014-22-1-2740, Accenture, the Stanford Institute for Human-Centered AI (HAI), and the Magic Grant from the Brown Institute for Media Innovation.

{
    \small
    \bibliographystyle{ieeenat_fullname}
    \bibliography{main}
}

\clearpage
\renewcommand\thefigure{S\arabic{figure}}
\setcounter{figure}{0}
\renewcommand\thetable{S\arabic{table}}
\setcounter{table}{0}
\renewcommand\theequation{S\arabic{equation}}
\setcounter{equation}{0}
\pagenumbering{arabic}%
\renewcommand*{\thepage}{S\arabic{page}}
\setcounter{footnote}{0}
\setcounter{page}{1}
\maketitlesupplementary
\appendix

\section{Dense View Generation}

Our goal is to generate dense, wide-range views of the underlying scene with the objects of interest from a single input image. 
The key here is to generate dense surrounding views with the scene content in the center and leverage the existing 3D reconstruction pipeline to reconstruct the underlying 3D scene. This approach is fundamentally different from the original image to 3D scene pipeline from WonderPlay~\cite{Wonderplay}. The previous pipeline relies heavily on the single-view depth estimation and the alignment between the original image and the inpainted regions, resulting in a 3D scene representation that can only be viewed in a narrow baseline. On the contrary, our current approach leads to a full, complete 3D scene that supports rendering from arbitrary viewpoints.

For dense view generation, we employ GEN3C~\cite{Gen3c}, which is capable of generating 3D-consistent multi-view videos from a single image. GEN3C is trained to generate multi-view videos of the underlying static scene, making it suitable for scene initialization. 
However, the vanilla GEN3C requires the generation to start from the input view. This presents a new challenge: generating a single, wide-angle~($180^{\circ}$ viewpoint change) camera trajectory often leads to inconsistent artifacts as the view deviates too far from the source~(\ie input image). 

To acquire a dense set of novel views while maintaining consistency, we split the required camera viewpoint changes into two separate trajectories: ``arc left" and ``arc right". Both trajectories originate from the input view and rotate in different directions, with a $90^{\circ}$ viewpoint change for each trajectory. We then generate a video for each trajectory and aggregate all the frames, forming a final, wide and consistent set of partial orbital views for the scene.

These generated dense views are further leveraged by our scene reconstruction pipeline for underlying 3D scene reconstruction, and also combined with SAM2~\cite{sam2} and Gaussian Grouping~\cite{gaussian_grouping} for foreground object segmentation.

\section{Object Mesh Generation}

Our pipeline further exploits TSDFusion~\cite{tsdfusion} to reconstruct the mesh for the foreground objects. However, we empirically find that this approach is suitable for highly deformable materials like fluid and granular objects, but for rigid objects, the texture and geometry artifacts from incomplete TSDFusion reconstruction can pose severe challenges for the video generation model and lead to unexpected hallucination during the video refinement process.

To enhance the robustness of meshes for rigid objects, we exploit additional steps to improve the reconstruction quality for objects of this material. 
Specifically, we introduce Hunyuan3D~\cite{hunyuan3d}, a powerful object-level image to 3D model. We directly leverage Hunyuan3D to generate the object mesh for rigid body objects, resulting in a complete surface mesh with much better quality compared to simple TSDFusion reconstruction, especially in the back regions. 

Unlike the non-robust depth alignment and manual object placement steps in WonderPlay, we can further benefit from our aforementioned dense view generation pipeline and automatically position this generated mesh into the scene. Leveraging our synthesized multi-view images, we optimize the $6$-DoF pose and scale by minimizing the projection error with respect to the surrounding views, ensuring it is perfectly aligned within the 3D scene.

\section{Physical Simulation Parameters}
The forward physics pass employs various solvers, and each solver requires specific physical parameter settings for reasonable simulation~\cite{Genesis}. We provide a comprehensive list of these parameters, along with their default values, in Table~\ref{tab:sim_params}. In practice, following the common practice in WonderPlay, these parameters are initially estimated using a Vision-Language Model~\cite{gpt4o} and are subject to optional manual fine-tuning to ensure physically plausible simulation results.

\section{Evaluation metric details}
To assess the scene quality, we render all generated scenes along a predefined camera trajectory and evaluate them using metrics from WorldScore~\cite{worldscore}. Specifically, we use rule-based metrics to validate camera controllability and 3D consistency. We also include the imaging metric to assess general per-frame visual quality. 
To evaluate the plausible physical dynamics, we conducted a user study with 350 participants for this aspect. We employed a Two-alternative Forced Choice~(2AFC) protocol, asking each participant to evaluate 10 scenes. For each scene, participants were given a multi-step interaction description and viewed a side-by-side, randomly ordered video comparison of our method and a baseline. Participants selected the video that performed better on one of two criteria: physics plausibility (the correctness of the predicted motion in response to the action) and motion fidelity (the quality and naturalness of the generated motion).

\section{Visual-physical aligned Particle Configuration}
While our VPP representation generally supports binding multiple gaussian primitives to a single physics particle, forming a set of size $K$, in practice, we configure $K$ and the gaussian scale adaptively based on material properties to ensure optimal visual-physical alignment:

\begin{itemize}
    \item \textbf{Solid and Surface Materials~(Rigid body, Cloth):} We set $K=1$. In this configuration, the gaussian scale is initialized to match the particle size $\delta$. This strict one-to-one mapping ensures that the visual appearance is tightly constrained by the physics simulation, effectively preventing visual artifacts such as ghosting or detachment during large deformations.
    
    \item \textbf{Volumetric and Emitter Materials~(Gas, Liquid, Sand, Snow, Elastic object):} To adequately represent the volumetric expansion and semi-transparent nature of these materials, we set $K=20$ to allow a single physics particle to cover a larger visual volume. Correspondingly, the VPP's gaussian scale is initialized to be smaller than the particle size, defaulting to $0.5\delta$, to represent fine-grained volumetric details within the particle's influence radius.
\end{itemize}

\section{Isotropic Visual Primitives}
\begin{figure}[t]
    \vspace{-0.5em}
  \centering
  \includegraphics[width=1.0\linewidth]{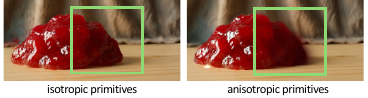}
  \vspace{-2em}
  
  \caption{Comparison of isotropic and anisotropic primitives in novel view synthesis.}
  \label{fig:rebuttal_isotropic}
\end{figure}

We demonstrate the differences between isotropic and anisotropic visual primitives in Figure~\ref{fig:rebuttal_isotropic}. We find that isotropic primitives help remove blurry artifacts in novel views, as they do not tend to overfit the input image.

\section{Ablation on Radius}
\begin{figure}[t]
  \centering
\vspace{-0.8em}
  \includegraphics[width=1.0\linewidth]{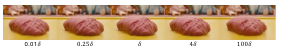}
  \vspace{-2.0em}
  \caption{Ablation on radius.}
  \label{fig:rebuttal_radius}
\end{figure}

We show an ablation on particle radius in Figure~\ref{fig:rebuttal_radius}. The default radius is set to $\delta$. Within a reasonable range (from $0.25\delta$ to $4\delta$), our results are robust to different values. However, an overly small radius ($\leq 0.01\delta$) leads to insufficient representational capability, and an overly large radius ($\geq 100\delta$) leads to instability in optimization.

\section{Limitation Discussion and Failure Case.} 
\begin{figure}[t]
  \includegraphics[width=1.0\linewidth]{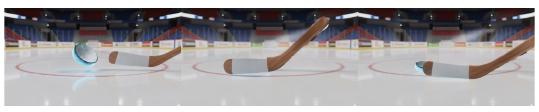}
  \vspace{-2em}
  \caption{Failure case in generating unseen geometry.}
  \label{fig:failure_case}
\end{figure}

We provide the detailed runtime breakdown in Table~\ref{tab:runtime_horiz}. \ourmethod is currently not real-time due to the backward optimization overhead. Figure~\ref{fig:failure_case} shows a failure case involving a hockey stick moving into the frame from out-of-view. In the middle, as the stick enters the field of view, it appears incomplete (a hockey stick should ideally be longer than it appears). It remains a future work to complete object geometry that is not seen in the input image.

\begin{table}[t]
    \centering
    
    \renewcommand{\arraystretch}{1.1} 
    \begin{tabular}{lc}
        \toprule
        Parameter & Default Value \\
        \midrule
        \textbf{General simulation} & \\
        Step time & $1e^{-3}$\\
        Sub-steps number & 10 \\
        Sampled particle size & $1e^{-2}$\\
        Gravity & $(0, 0, -9.8)$ \\
        \midrule
        \textbf{Rigid body solver} & \\
        friction coefficient & 0.1 \\
        \midrule
        \textbf{MPM solver} & \\
        Grid density & 64\\
        Elastic material Young's modulus & $3e^5$ \\
        Elastic material Poisson's ratio & 0.2 \\
        Liquid material Young's modulus & $1e^7$ \\
        Liquid material Poisson's ratio & 0.2 \\
        Granular material Young's modulus & $1e^6$ \\
        Granular material Poisson's ratio & 0.2 \\
        Granular material Friction angle & 45 \\
        \midrule
        \textbf{PBD solver} & \\
        Cloth material stretch compliance & $1e^{-7}$ \\
        Cloth material bending compliance & $1e^{-5}$ \\
        Smoke material viscosity coefficient & 0.1 \\
        \bottomrule
    \end{tabular}
    \caption{Simulation parameters and default values}
    \label{tab:sim_params}
\end{table}
\begin{table}[t]
\centering
\resizebox{1.0\linewidth}{!}{
\begin{tabular}{l|ccc|c}
\toprule
\textbf{Stage} & \textbf{Initialization} & \textbf{Forward Pass} & \textbf{Backward Opt.} & \textbf{Total (1st Loop)} \\
\midrule
\textbf{Time} & $\sim$8 min & $<$1 min & $\sim$7 min & $\sim$16 min \\
\bottomrule
\end{tabular}
}
\vspace{-0.5em}
\caption{Runtime Analysis.}
\label{tab:runtime_horiz}
\end{table}

\end{document}